# Design and Optimization of Reinforcement Learning-Based Agents in Text-Based Games

Haonan Wang [1,*], Mingjia Zhao [2], Junfeng Sun [2], Wei Liu [2]

[1] Johns Hopkins University, Baltimore, USA

[2] College of Science, Liaoning Technical University, China

* Corresponding Author: Haonan Wang (Email: hwang298@jh.edu)

**ABSTRACT**

As AI technology advances, research in playing text-based games with agents has become progressively popular. In this paper, a novel approach to agent design and agent learning is presented with the context of reinforcement learning. A model of deep learning is first applied to process game text and build a world model. Next, the agent is learned through a policy gradient-based deep reinforcement learning method to facilitate conversion from state value to optimal policy. The enhanced agent works better in several text-based game experiments and significantly surpasses previous agents on game completion ratio and win rate. Our study introduces novel understanding and empirical ground for using reinforcement learning for text games and sets the stage for developing and optimizing reinforcement learning agents for more general domains and problems.

## 1. INTRODUCTION

As AI evolves, its potential to manage complex challenges and sophisticated tasks is more apparent. As a unique style of game-playing, text-based games require high language understanding and logical reasoning capabilities, making it a primary medium to determine AI capability. However, it is hard to properly optimize decision-making in text-based game-playing agents and derive useful knowledge from large pools of game text. Reinforcement learning, a process of optimization of decision strategies using trial-and-error learning and environmental feedback, has been a success in a large spectrum of tasks, such as video games and board games. The use of reinforcement learning in designing and optimization of agents has been widely investigated. However, its use in text-based games, in particular in deep reinforcement learning, is a promising avenue to be investigated. The work in this paper is interested in designing and optimizing reinforcement learning-based agents in text-based games to manage optimization of agent decision-making challenges. The work includes designing deep learning models to process game text and using policy gradient-based deep reinforcement learning in learning agents. The aim is to optimize the capability of agents in text-based games to facilitate AI advancement.

## 2. AI TECHNIQUES IN TEXT-BASED GAMES

### 2.1. Overview of AI Applications in Text-Based Games

Text-based games employ natural language to interact, in which players input text to interact in a game world. The games are applicable in AI studies due to their high action freedom and sophisticated scene description. AI in text-based games is more concerned with natural language processing (NLP) approaches to parse players' input commands and parse stories in games to enable semantically reasonable interactions. Decision making and planning capabilities are also introduced to enable agents to reason in sophisticated dynamic text worlds to generate strategies. Generative and comprehension models in current work assist in enabling more efficient text parsing and decision capabilities in agents. The sophisticated language worlds of text-based games combined with their strategic challenges of a long-term nature provide a principal proving ground for reinforcement learning to manage NLP. The combination has allowed AI to handle better task complexity and adaptability of the environment.

### 2.2. The Role of Reinforcement Learning in Text-Based Games

Reinforcement learning is of great use in text-based games in that it facilitates learning of strategies of agents in a process of playing a game environment. The problem of text-based games is that they use text to deliver information, making their state spaces high in dimensionality, high in level, and requiring high reasoning and decision making capability of their agents. Reinforcement learning, in a trial-and-error process in a dynamic process of receiving a response, facilitates learning of optimal strategies in complex game tasks in a process of forming efficient behavioral patterns. Reinforcement learning's reward system facilitates dynamic adjustment of strategies to a completed task, iteratively improving their behavior to a point of optimization. The application of reinforcement learning has introduced new means of applying AI to text-based games, making it more efficient to design agents.

### 2.3. Limitations of Traditional Agent Design

Traditional agent architectures in computer games use rule-based or template-based methods that fail to provide generalization and cannot cope with dynamic and complex circumstances in a game. With a large amount of game text to be encountered, such agents do not reason globally to produce poor and suboptimal action options. Classical architectures also fail to cope with dynamically changing state spaces, rendering it challenging to apply them to more complex tasks.

## 3. IMPLEMENTATION OF DEEP LEARNING MODELS IN TEXT-BASED GAMES

### 3.1. Parsing Game Text and Constructing World Models

In text-based games, there is a need to parse text description to derive an environmental perception, putting heavy loads on modeling and parsing abilities of a model. Deep learning models can parse game text to derive implicit environmental structure and informative facts, accurately rebuild a game's world state. Pre-trained language models, such as learned via the Transformer structure, manage contextual dependencies and semantic relationships well, making it helpful in recognizing goals, task tips, and interactive objects of an environment in a game scene. From parse results, a world model is set up to reason over rules and logics of a game. The world model causally links potential state spaces in a game to forecast potential resulting state transitions of agent action. The structured modeling approach achieves a stable environmental representation for reinforcement learning agents, providing a foundation for optimization of strategies that follows.

## 3.2. Training Agents Using Deep Learning

Deep learning is utilized to train text-based game-playing agents to enhance their decision making and comprehension abilities. By defining neural networks, text in a game is utilized to parse semantics and construct representations of a state. Parsed text is utilized to input a deep learning model during training that is conditioned using supervised learning or generation models to derive implicit text relationships to construct usable state features that input to reinforcement learning to provide a platform to learn a policy. Training is ensured to provide model generalizability using high-quality training set and multi-task learning approaches to enable better adaptability in different text scenarios. The effectiveness of training of the agent is measured using end metrics such as text parsing accuracy, coherency of policy execution, and achievement of game objectives, providing considerable improvement in text comprehension and decision making.

## 3.3. Transforming State Value into Optimal Policy

The transformation of optimal policy is accomplished via policy gradient-based deep reinforcement learning. Deep learning models in text-based game settings extract features that the agents apply to estimate game states in a quantitative aspect to produce state value functions. In learning, optimization of policy iteration enables the closure of gaps between optimal decisions and state value to produce maximization of rewards. The method reduces action space exploration cost and maximizes policy convergence efficiency to deliver a stable support to agents' performance in complex task situations.

## 4. POLICY GRADIENT-BASED DEEP REINFORCEMENT LEARNING METHODS

### 4.1. Overview of Policy Gradient Methods

Policy gradient methods represent a basic method in deep reinforcement learning that models and maximizes policies to support efficient and adaptive decision making. Most methods of reinforcement learning would involve building and updating value functions in traditional methods, yet in high-dimensioned state or complex tasks, such methods would not converge or be costly in calculation. Policy gradient methods suggest parameterized policy functions that maximise cumulative return in expectation without using value functions. The fundamental nature of such a method is based on the reinforcement learning objective function, using gradient ascent methods to iteratively adjust policy parameters to enable learning more optimal decision strategies in the agent. In practical use, policy gradient methods estimate the objective function gradient using sampled trajectory data to alleviate challenges introduced by high-dimensioned state spaces. As a basic branch of deep reinforcement learning, policy gradient methods dramatically enhance adaptability and performance in reinforcement learning applications.

### 4.2. Implementing Policy Gradient Methods in Agent Training

In agent learning, fundamental optimization tools in text-based games apply policy gradient approaches to improve decision making capabilities. Deep neural networks defining the policy function determine action based on a distribution of the policy function in a state. With Monte Carlo or temporal difference approaches combined, optimization goals of resulting returns of the agent in a game are estimated. With maximization of cumulative cumulative rewards, policy gradient approaches adjust network parameters using gradient ascent, iteratively updating the behavior policy of the agent. Advantage functions are used to manage variance to improve learning efficiency to guide better optimization in improving the policy gradient. With such a process, agents dynamically adjust strategies in advanced text-based game situations to better support diverse demands of tasks.

### 4.3. Advantages of Policy Gradient-Based Deep Reinforcement Learning Over Traditional Methods

Policy gradient-based deep reinforcement learning is highly robust in text-based video games over rule-based or model-based agents. The method maximizes policies directly, making it more easily probable to discover better problem solutions in high-dimensional action spaces, enhancing action flexibility in making a decision. Policy gradient methods learn to adapt to dynamic game scenarios without high-dimensional state space requirements in computations. Deep learning models combined facilitate more efficient game text comprehension in the agent to better carry out missions at higher ratios combined with more efficient generalization in complex missions.

## 5. OPTIMIZING AGENT PERFORMANCE IN TEXT-BASED GAMES

### 5.1. Design Methods for Optimizing Agents

Optimizing agent design is dedicated to enhancing adaptability and decision making in text-based games. With modular design, the agent is separated into function modules of text parsing, action generation, and adjustment of feedback. The text parsing module, developed based on pre-trained language models, accurately analyzes semantic knowledge in game text, enabling dynamic updating of the world model. The action generation module, developed based on policy gradient methods, facilitates efficient and accurate generation of strategies in a complex world. The adjustment module of feedback adjusts the decision-making model in response to rewards in the game, enabling dynamic adjustment of strategies in a flexible way to adapt to dynamic changes in the environment. In order to further enhance agent performance, mechanisms of experience replay and prioritized sampling methods are utilized to stabilize and quicken the process of training, preventing overfitting using regularization methods. All of these design approaches greatly enhance agent performance in a large number of text-based game tasks, laying a foundation for efficient completion of tasks.

### 5.2. Performance of Optimized Agents in Text-Based Game Tests

Optimized agents gain more in various text-based game experiments. In task completion, optimized agents are more efficient and accurate compared to regular agents, suggesting better perception of the environment and decision-making abilities. In win ratios, in multiple rounds of game experiments, optimized agents obviously surpass regular agents, suggesting their superiority in complex game scenarios. In new game challenges, optimized agents demonstrate stronger generalization abilities, learning new rules of a game or a scene easily. The experimental results also indicate that reinforcement learning agents created using certain optimization methods can adapt to challenges in text-based games well, providing a technical support to related applications.

### 5.3. Comparison Between Optimized Agents and Previous Agents

Optimized agents also outdo traditional agents in optimization in text-based game experiments. Optimized agents get better winning ratios and game completion ratios in different experiments, suggesting better capabilities of task parsing and decision making. The traditional agents also fail in dynamic adjustment of strategies in dynamic scenarios, but optimized agents, based on deep reinforcement learning, have high adaptability in the environment and optimal adjustment capabilities of strategies, rendering them useful and technically superior in text-based games.

# 6. FUTURE APPLICATIONS OF REINFORCEMENT LEARNING IN TEXT-BASED GAMES

## 6.1. Potential Applications in More Complex Environments and Tasks

The potential of reinforcement learning in text-based games is transitioning to more advanced contexts and applications. Text-based games often involve large state spaces, incomplete knowledge, high-dimension natural language interactions, which also occur in a great number of real-world applications, such as dialogue systems, automatic puzzle-solving, natural language-based execution of tasks. By advancing reinforcement learning agents in text-based games, useful knowledge can be transferred to more advanced real-world applications. Reinforcement learning can be integrated with more advanced NLP algorithms, such that contextual semantics can be learned, reasoning can be done, and strategies can be produced more accurately. In the future, it is possible to apply cooperation of multiple agents in text-based game situations, to explore collective intelligence in higher-dimension interactive situations. With advancing reinforcement learning and deep learning technologies, their applications in more advanced text-based games and related applications can extend beyond classic one-step decision problem to involve long-term planning, dynamic adjustment, and fusion of multiple modes of information in higher-level intelligent decision situations.

## 6.2. Future Trends in Agent Design and Optimization

Future trends in designing and developing agents involve more environmental adaptability, more efficient learning mechanisms, and more generalized tasks. In environmental adaptability, agents must be able to handle more dynamic and more complex text-based game situations, requiring stronger reinforcement learning models of more strength of generalization and more realistic simulated situations. Efficient learning mechanisms will move more towards more advanced deep reinforcement learning approaches, i.e., self-supervised learning and multi-modal learning, to facilitate higher learning speed and decision making accuracy. In task generalization, future agents must be able to cross over across domains, easily switching between different text-based games or different rule-based tasks, to facilitate multi-task processing and cross-domain cooperation. The trend will facilitate incessant optimization of agent capability in text-based games and provide technical support for their application in other complex task situations.

## 6.3. Recommendations for Optimizing Reinforcement Learning Applications Based on This Study

Optimizing reinforcement learning agents is possible via more efficient processing of advanced semantics, improving their adaptability in terms of policy, and their generalizability in new, unobserved contexts, hence tackling more demanding text-based applications.

# 7. CONCLUSION

This study discusses design and optimization of text-based game agents in a new method using reinforcement learning. Deep learning models process game text to construct up world models, to be subsequently followed up by reinforcement learning based on policy gradient to learn agents, to allow it to transform state value to optimal policy. The experimental results show that this approach of optimization has high completion ratios of text-based games and win ratios over existing agents to a large extent. However, although promising, design and optimization of reinforcement learning agents in more realistic situations of text-based games and scenarios is a fundamental and hard problem that must be investigated more profoundly. There are also constraints in this work, such as open questions in training effectiveness and adaptability to new scenarios. More studies must delve more in design and optimization of agents in learning and decision making to allow more improvement in agent performance in various text-based games. Overall, this work introduces new knowledge and practical ground to apply reinforcement learning to text-based games, to provide a solid background to more studies to be established in the future.